\documentclass[letterpaper]{article} 
\pdfoutput=1

\usepackage{aaai25}  
\usepackage{times}  
\usepackage{helvet}  
\usepackage{courier}  
\usepackage[hyphens]{url}  
\usepackage{graphicx} 
\usepackage{natbib}  
\usepackage{caption} 
\usepackage{subfig}
\usepackage{amsmath,amsfonts}
\usepackage{amssymb}
\usepackage{array}
\usepackage{multirow}
\usepackage[table,xcdraw]{xcolor}

\usepackage[capitalize]{cleveref}
\crefname{section}{Sec.}{Secs.}
\Crefname{section}{Section}{Sections}
\crefname{table}{Tab.}{Tabs.}
\Crefname{table}{Table}{Tables}

\usepackage{xspace}
\makeatletter
\DeclareRobustCommand\onedot{\futurelet\@let@token\@onedot}
\def\@onedot{\ifx\@let@token.\else.\null\fi\xspace}

\def\eg{\emph{e.g}\onedot} 

\def\ie{\emph{i.e}\onedot}

\title{Modality-Aware Shot Relating and Comparing for Video Scene Detection}
\author{
    Jiawei Tan\textsuperscript{\rm 1, \rm 2},
    Hongxing Wang\textsuperscript{\rm 1, \rm 2}\thanks{Corresponding author: Hongxing Wang.},
    Kang Dang\textsuperscript{\rm 3},
    Jiaxin Li\textsuperscript{\rm 1, \rm 2},
    Zhilong Ou\textsuperscript{\rm 1, \rm 2}
}
\affiliations{
    \textsuperscript{\rm 1}Key Laboratory of Dependable Service Computing in Cyber Physical Society (Chongqing University), Ministry of Education, China\\
    \textsuperscript{\rm 2}School of Big Data and Software Engineering, Chongqing University, China\\
    \textsuperscript{\rm 3}School of AI and Advanced Computing, XJTLU Entrepreneur College (Taicang), Xi’an Jiaotong-Liverpool University, Suzhou, China\\
    \{jwtan, ihwang\}@cqu.edu.cn, Kang.Dang@xjtlu.edu.cn, jiaxin\_li@cqu.edu.cn, zlou@stu.edu.cn
}

\begin{document}

\maketitle

\begin{abstract}
Video scene detection involves assessing whether each shot and its surroundings belong to the same scene.
Achieving this requires meticulously correlating multi-modal cues, \eg visual entity and place modalities, among shots and comparing semantic changes around each shot.
However, most methods treat multi-modal semantics equally and do not examine contextual differences between the two sides of a shot, leading to sub-optimal detection performance.
In this paper, we propose the \textbf{M}odality-\textbf{A}ware \textbf{S}hot \textbf{R}elating and \textbf{C}omparing approach (MASRC), which enables relating shots per their own characteristics of visual entity and place modalities, as well as comparing multi-shots similarities to have scene changes explicitly encoded. Specifically, 
to fully harness the potential of visual entity and place modalities in modeling shot relations, we mine long-term shot correlations from entity semantics while simultaneously revealing short-term shot correlations from place semantics.
In this way, we can learn distinctive shot features that consolidate coherence within scenes and amplify distinguishability across scenes.
Once equipped with distinctive shot features, we further encode the relations between preceding and succeeding shots of each target shot by similarity convolution, aiding in the identification of scene ending shots.
We validate the broad applicability of the proposed components in MASRC. Extensive experimental results on public benchmark datasets demonstrate that the proposed MASRC significantly advances video scene detection.
\end{abstract}
\begin{links}
\link{Code}{https://github.com/ExMorgan-Alter/MASRC}
\end{links}

\section{Introduction}
Video scene detection involves identifying whether a shot is at the scene ending or not. Upon detection, the video is segmented into coherent sets of scenes. Temporal scenes provide structured information that serves as the foundation for downstream applications such as text-to-video retrieval~\cite{scene_app1} and human-centric storyline construction~\cite{scene_app2}.

\begin{figure}[!t]
	\centering
	\subfloat[Relating]{\label{fig:problem}\includegraphics[width=\linewidth]{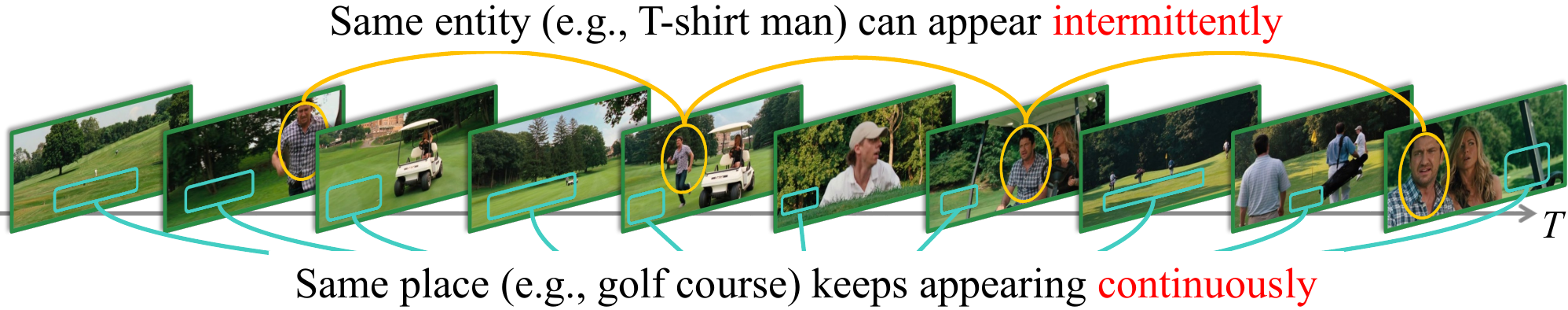}}
{}
	\subfloat[Comparing]{\label{fig:problem2}\includegraphics[width=\linewidth]{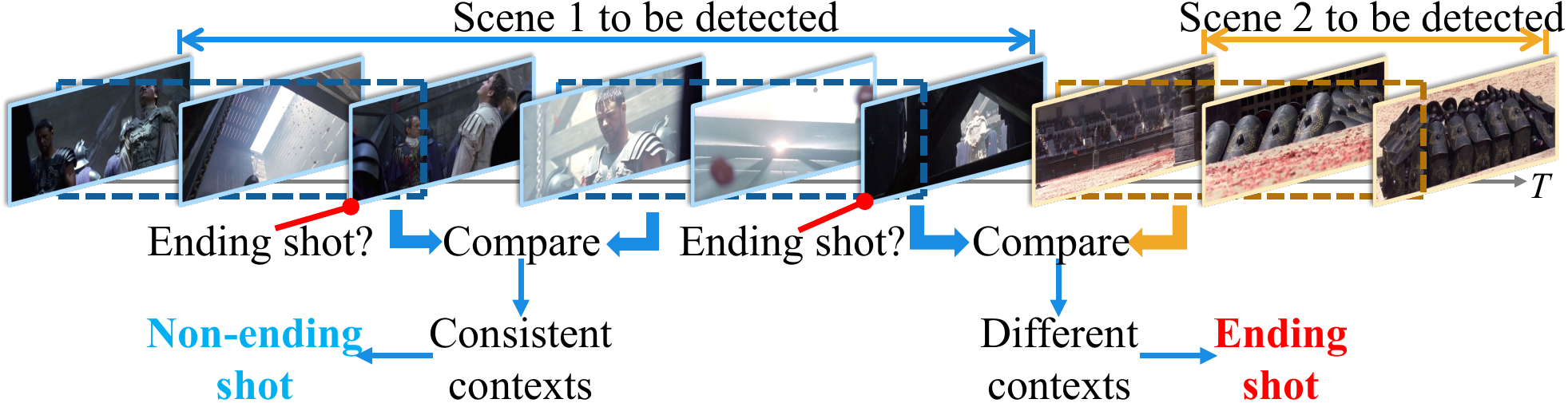}}
	\caption{Intuition behind our proposed modality-aware shot relating and comparing. (a) Relating: We capture long-term shot relations by linking intermittently appearing identical entities and short-term shot relations by relating consecutive shots depicting the same place. (b) Comparing: Consistency drawn from comparisons between fore-and-aft contexts is an indicator to classify whether a target shot is an ending shot.}
\end{figure}

\begin{figure*}[!t]
	\centering
	\includegraphics[width=.99\linewidth]{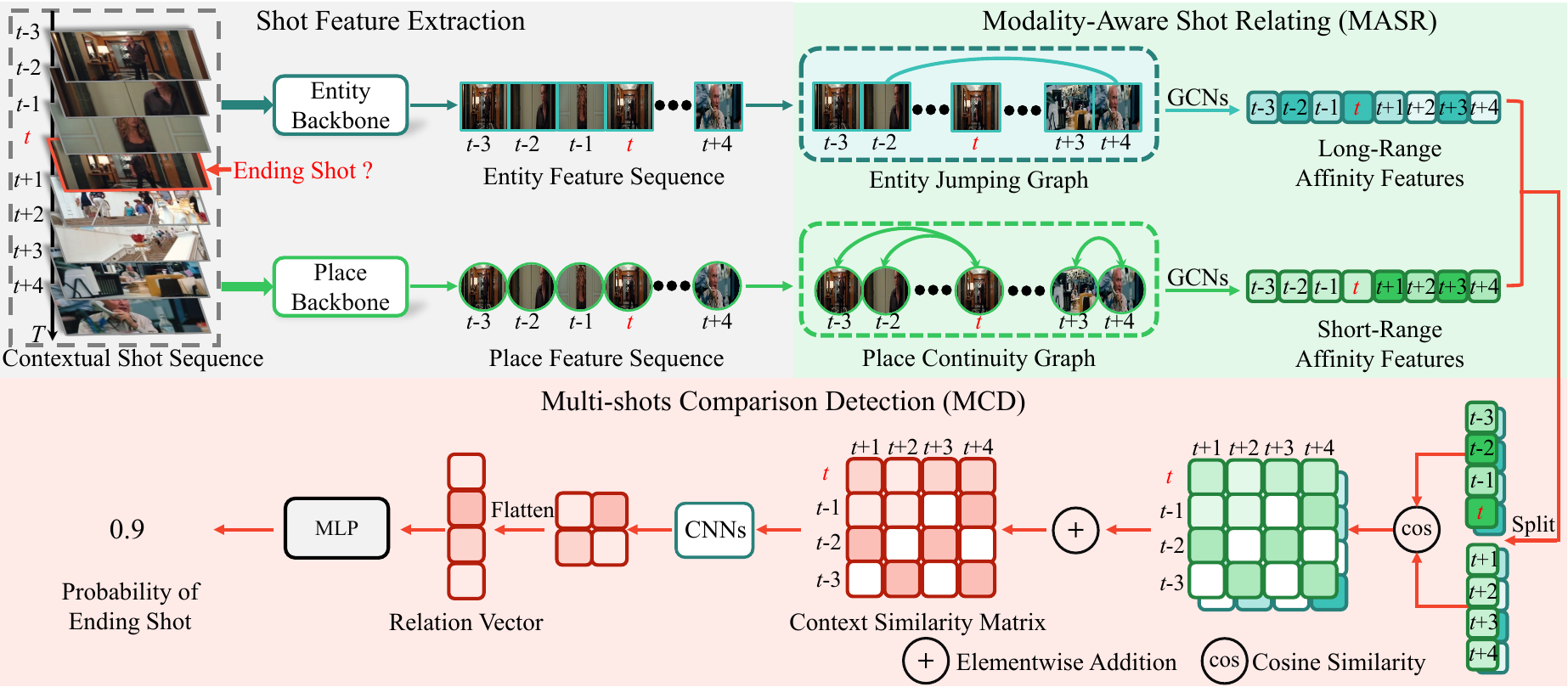}
	\caption{Diagram of how the proposed MASRC determines whether a target shot (in the red border) is an ending shot or not. We build an entity jumping graph and a place continuity graph for GCN message passing to separately embed long and short shot relations into shot representations. Relying on similarity comparison between the fore-and-aft shots of each target shot as well as further similarity change encoding by convolution, the probability of target shots being ending shots can be better predicted by a simple MLP classifier.}
	\label{fig:pipeline}
\end{figure*}

Before determining whether a shot is an ending shot, one needs to examine the surrounding shots to grasp the context of the target shot from different visual modalities, especially in terms of imaging entities and places, in the video.
This process requires associating shared semantics between shots for each modality to model shot relations.
As shown in \cref{fig:problem}, we need to intermittently link shots featuring the same entity for capturing long-term shot relations, while relating consecutive shots depicting the same place for the short-term shot relations.
Common methods~\shortcite{msd_lgss,msd_mhrt,msd_vsmbd} handle different modalities in indiscriminate manner for shot relations modeling.
However, this modality-agnostic strategy fails to capture the diverse temporal relations exhibited by different modalities.
To address this issue, we shift temporal relation modeling from modality-agnostic recipes to modality-aware scheme.
As presented in \cref{fig:pipeline}, we design Modality-Aware Shot Relating (MASR) to enable capturing long-range and short-range temporal relations between shots respectively in entity and place modalities.
In the entity modality, we construct an Entity Jumping Graph (EJG) to correlate shots with similar entities by which a long-term dependencies can be caught for similar but distant shots.
With the place modality, we design a Place Continuity Graph (PCG) to connect time-continuous shots depicting the same place, modeling short-term dependencies between shots.
%

Once modeling shot relations through MASR, we are allowed to classify whether each shot is an ending shot via its own context-embedded shot feature, as previous efforts~\cite{msd_scrl,msd_bassl,msd_tran4fer,msd_m2s,msd_cat} have done.
Unfortunately, a single context-embedded shot feature struggles to capture complex context changes, leading to numerous false positives.
This is because these methods overlook the fact that an ending shot signifies a contextual difference between its two sides.
As illustrated in \cref{fig:problem2}, it is crucial to compare the consistency between fore-and-aft contexts using context-embedded shot features.
\cref{fig:pipeline} shows that devised Multi-shots Comparison Detection (MCD), which compares multi-shot similarities and then encodes similarity changes by convolution, aiding the identification of ending shots.

Across the board, we propose a novel architecture for video scene detection, dubbed \textbf{M}odality-\textbf{A}ware \textbf{S}hot \textbf{R}elating and \textbf{C}omparing (MASRC). 
We pursue MASR to capture diverse shot relations for entity and place modalities. Next, we design MCD to encode contextual different between shots to support the classification of ending shots.

In a nutshell, our contributions include:
\begin{itemize}
\item We propose MASR to capture long-range and short-range shot relations by considering the distinctive roles of entity and place modalities in video scene detection.
\item Instead of relying solely on individual shots for video scene detection, we compare shot semantics on both sides of the target shot and propose MCD to better encode and detect scene transitions.
\item We perform comprehensive evaluations on three video scene detection datasets including MovieNet~\cite{msd_movienet}, BBC~\cite{msd_bbc}, and OVSD \cite{msd_ovsd}.
Our proposed method surpasses previous approaches by large margins in various learning settings.
\end{itemize}

\section{Related Work}
\textbf{Video Scene Detection.}
Early methods directly employ shot features to cluster adjacent shots into scenes in an unsupervised manner.
For instance, \cite{msd_sshv} clusters shots based color histograms and temporal closeness by a spectral clustering method.
Similarly, \cite{msd_mbvr} employs color histogram intersection for initial shot clustering, followed by merging the preliminary results using a sliding time window to derive scenes.
\cite{msd_graphcut} utilizes color and motion information between shots to generate a similarity graph, partitioning the graph to identify video scenes.
However, these methods achieve limited performance due to their manually modeling relations between shots, resulting in insufficient discriminability of shot features from different scenes.
Recently, supervised methods focus on designing adaptive and flexible relating shot-correlation mechanisms.
Some methods exploit single-modality visual shot semantics to explore shot relations.
For instances, \cite{msd_tp} and \cite{msd_bassl} employ a multi-head attention mechanism to emphasize long-term shot relationships. \cite{msd_cat} additionally incorporates local window attention to capture short-term shot relationships.
\cite{msd_canet} reweights each shot feature to enhance the ability of LSTM~\cite{lstm} to correlate shots within long video scenes.
Due to the limitations of single modality in semantic representation, other methods employ multi-modality visual shot semantics, such as entity, place modalities.
\cite{msd_lgss} designs multi-modal shot boundary features for capturing both difference and relations between neighborhood shots.
\cite{msd_mhrt} employs a multi-head self-attention mechanism on the similarity matrix generated by multi-modal shot features for exploring high-order relationships between shots.
However, these methods tend to design a unified framework to handle different modalities, struggling to capture the diverse temporal relations exhibited by different modalities.
In contrast, we propose MASR to captures long-range and short-range shot relations by considering the distinctive roles of entity and place modalities in video scene detection.

On the other hand, when detecting ending shots, some methods~\cite{msd_scrl,msd_bassl,msd_tran4fer,msd_m2s} make a prediction by a single context-embeded shot representation.
Other methods~\cite{msd_mhrt, msd_tp} design learnable boundary class vectors and use it to query each shot feature to detect whether each shot is an ending shot.
However, these methods overlook the fact that an ending shot signifies a semantic difference between its two sides and do not explicitly compare the shot contexts for detection.
In this paper, we compare the shot semantics on both sides of the target shot to identify the ending shots.

\textbf{Graph Neural Network.}
In recent years, an increasing number of non-Euclidean data have been represented as graphs, posing significant challenges to existing neural network methods. To effectively deal with such data, Graph Neural Networks (GNNs) have attracted much attention. GNNs have been applied in various fields, including recommendation systems~\cite{gcn_recom}, computer vision~\cite{gcn_cv}, and natural language processing~\cite{gcn_nlp}. Graph convolutional networks, a type of GNN, can be divided into spectral-based and spatial-based approaches~\cite{gcn_survey}. Spectral-based approaches implement graph convolution by defining filters similar to graph signal processing, while spatial-based approaches define graph convolution by information propagation and have gained momentum for their efficiency, flexibility, and generality. Widely used GNN techniques include Graph-SAGE~\cite{gcn_sage}, GAT~\cite{gcn_gat}, and GCN~\cite{gcn}.
\section{Methodology}
\subsection{Problem Formulation}
Given a video divided into shots, video scene detection aims to learn a classifier $f$, which holds $f(s) = 1$ if shot $s$ ends a scene, $f(s) = 0$ otherwise.
Since a shot itself cannot form the concept of the end or non-end of a scene, we have to put a shot in its temporal context towards this end.
To be explicit, we rewrite $f(s)$ as $f(c(s))$, where $c(\cdot)$ truncates the contextual sequence of shots including $s$.
In this study, we propose a Modality-Aware Shot Relating and Comparing solution (MASRC). As illustrated in \cref{fig:pipeline}, it allows Modality-Aware Shot Relating (MASR) for Multi-shot Comparison Detection (MCD) to model $f(c(s))$.

\subsection{Modality-Aware Shot Relating}
\label{subsec:mldm}
To perform prediction on shot $s_t$ at time $t$, we cut its context $c(s_t)$ as a time sliding window of length $T$ centered around $s_t$.
Since the frames within a shot are similar, we represent each shot with a randomly selected frame, as is common practice~\cite{msd_bassl, msd_cat, msd_neighbor}. 
Without loss of generality, we let $T$ be even, and denote $c(s_t)$ by $\{s_{t-{T/2}+1}, \cdots, s_t, \cdots, s_{t+{T/2}}\}$.
Given the importance of actors, objects, and places in composing visual-centric shot semantics~\cite{msd_lgss}, we directly employ different pre-trained ResNet-50 models~\cite{resnet} to extract diverse visual semantics, as done in~\cite{msd_vsmbd}.
One model, pre-trained on the ImageNet dataset~\cite{imagenet}, extracts visual entity features related to actors and general objects, \ie, ${{\boldsymbol{X}}^{\rm{E}}} = \{ {\boldsymbol{x}}_{t-{T/2}+1}^{\rm{E}},...,{\boldsymbol{x}}_{t+{T/2}}^{\rm{E}}\}$. Another ResNet-50 model, pre-trained on the Places dataset~\cite{place365}, is used to obtain place features ${{\boldsymbol{X}}^{\rm{P}}} = \{ {\boldsymbol{x}}_{t-{T/2}+1}^{\rm{P}},...,{\boldsymbol{x}}_{t+{T/2}}^{\rm{P}}\}$ for shot sequence $c(s_t)$.

\subsubsection{Entity-based Long-Term Dependency.}\label{subsubsec:ald}
The reappearance of the same actors or objects in a shot after several shots reflects that entities carry long-range shot relations.
To relate shots with similar entity semantics, we construct an entity jumping graph (EJG) ${G^{\rm{E}}} =  < {{\boldsymbol{X}}^{\rm{E}}},{{\boldsymbol{E}}^{\rm{E}}} >$.
The shots with features ${{\boldsymbol{X}}^{\rm{E}}}$ act as nodes and edges ${{\boldsymbol{E}}^{\rm{E}}} = \{E_{ij}^{\rm{E}} \}$ between nodes are determined by the semantic similarity between shots, given by:
\begin{equation}
E_{ij}^{\rm{E}} = \left\{ {\begin{array}{*{20}{l}}
{S_{ij}^{\rm{E}},}&{S_{ij}^{\rm{E}}\; \in {\rm{top}}{\textit{-k}}\;({\bf{S}}_i^{\rm{E}}),}\\
{0,}&{{\rm{otherwise}},}
\end{array}} \right.
\label{eq:vg}
\end{equation}
where cosine similarity ${\rm{S}}_{ij}^{\rm{E}} = \rm{cos}({{\boldsymbol{x}}^{\rm{E}}_i}, { {\boldsymbol{x}}^{\rm{E}}_j})$ is computed based on entity features ${{\boldsymbol{x}}^{\rm{E}}_i}$ and ${{\boldsymbol{x}}^{\rm{E}}_j}$. EJG establishes connections between each shot and its $k$ most similar shots in the entity feature space, enabling each shot to be linked with distant but similar shots in the temporal dimension.

After that, we perform message passing for the shot nodes on ${G^{\rm{E}}}$ with shot features ${{\boldsymbol{X}}^{\rm{E}}}$.
In this process, the shot nodes will receive messages from its connected shots in ${G^{\rm{E}}}$, facilitating information exchange between distant shots.
Specifically, we stack two graph convolution network (GCN)~\cite{gcn} layers on EJG to model long-term shot relations for obtaining long-range affinity features ${{\boldsymbol{X}}^{\rm{LR}}}$:
\begin{equation}
\left\{ \begin{array}{l}
{\boldsymbol{X}}_0^{{\rm{LR}}} = {{\boldsymbol{X}}^{\rm{E}}},\\
{\boldsymbol{X}}_i^{{\rm{LR}}} = {\rm{LN}}({\boldsymbol{X}}_{i - 1}^{{\rm{LR}}} + \sigma ({\rm{GC}}{{\rm{N}}_i}({\boldsymbol{X}}_{i - 1}^{{\rm{LR}}},{{{\boldsymbol{\tilde E}}}^{\rm{E}}}))),
\end{array} \right.
\label{eq:feat_lr}
\end{equation}
where, in each layer $i \in \{1, 2\}$, we apply vanilla GCN smoothing to current features ${\boldsymbol{X}}_{i - 1}^{{\rm{LR}}}$ on the edge affinities ${{\boldsymbol{\tilde E}}^{\rm{E}}}$ normalized from ${{\boldsymbol{E}}^{\rm{E}}}$ \cite{gcn}, followed by activation mapping with $\sigma(\cdot)$.
The resultant activated features will be added to original features ${\boldsymbol{X}}_{i - 1}^{{\rm{LR}}}$ as smoothly rectified features for layer normalization (LN) \cite{layer_norm}.

\subsubsection{Place-based Short-Term Dependency.}\label{subsubsec:psd}
\begin{figure}[!t]
\centering
	\includegraphics[width=.99\linewidth]{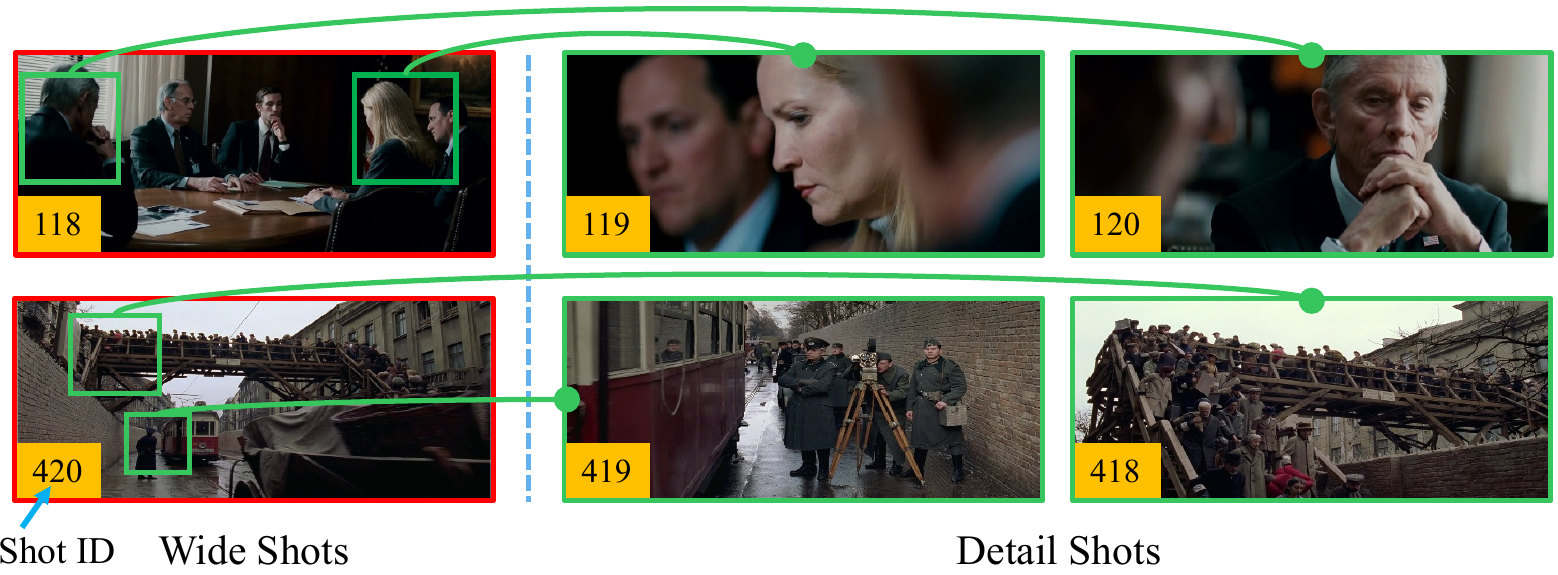}
	\caption{Each wide shot depicts an overview of a place, and its detail shots zoom in on specific details within the same place.}
	\label{fig:wide_detail}
\end{figure}

~Consecutive shots in the same scene often share the same place.
Establishing correlations among these shots allows us to effectively model short-range shot relations.
However, the challenge lies in grouping shots that depict the same place.
As shown in \cref{fig:wide_detail}, some shots provide an overview of the layout for place, while others focus on on specific details within the same place.
For clarity, we refer shots portraying the overall layout of the place to ``wide shots" and the remaining shots to ``detail shots".
That is to say, wide shot can have similarities to its detail shots, but detail shots are usually very diverse.
Hence, a wide shot has more similar shots than its detail shots.
With this observation, we count the number of similar shots to each involved shot to identify wide and detail shots.
\begin{figure}[!t]
\centering
	\includegraphics[width=.95\linewidth]{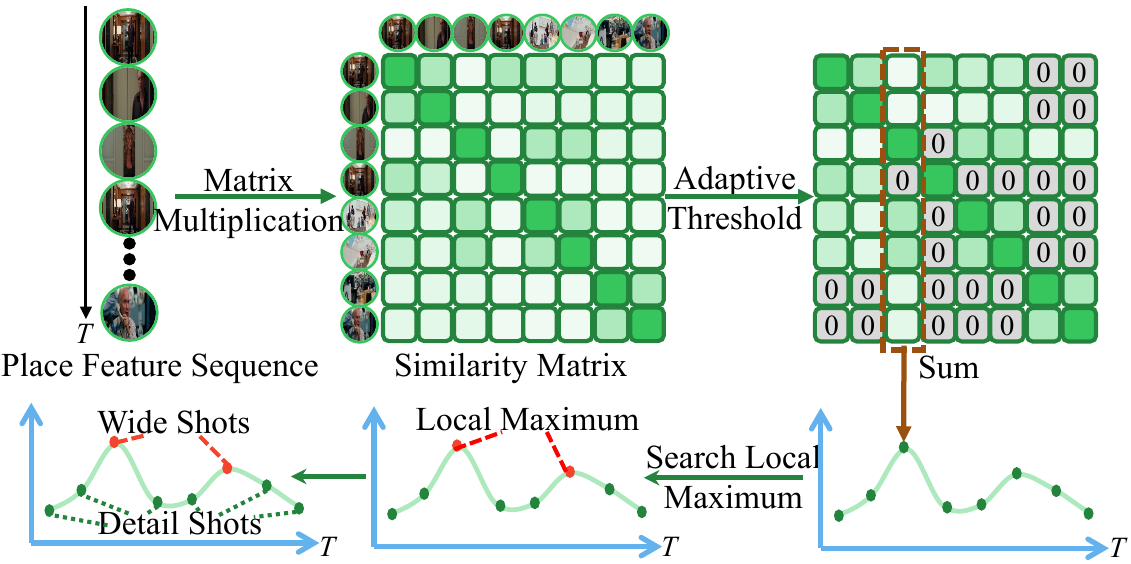}
	\caption{Identification of wide shots and detail shots.
}
	\label{fig:place_graph}
\end{figure}
As shown in \cref{fig:place_graph}, we leverage an adaptive threshold to obtain the number of similar shots to shot ${s_i}\in c(s_t)$ as follow,
\begin{equation}
{{n_i} = {\mathop{\sum} \limits_{j \in [{t - \frac{T}{{\rm{2}}} + 1,t + \frac{T}{{\rm{2}}}}]} }{{\rm I}({S_{ij}^{\rm{P}}} > {\bar S^{\rm{P}}} )}},
\label{eq:threhold}
\end{equation} 
where ${S_{ij}^{\rm{P}}} = \rm{cos}({{\boldsymbol{x}}^{\rm{P}}_i}, { {\boldsymbol{x}}^{\rm{P}}_j})$ denotes cosine similarity between features from the place modality, ${\bar S^{\rm{P}}}$ denotes the averaged similarity among shot features of place modality, and ${\rm{I}}( \cdot )$ refers to an indicator function that takes $1$ for non-zero inputs.
According to $\{ {n_i}\}_{i\in[{t-{T/2}+1},{t+{T/2}}]}$, we select those shots $\{s_\mathcal{W}\}$ with $\mathcal{W} = \{ i \in [{t-{T/2}+1},{t+{T/2}}]|{n_i} > {n_{i - 1}} \cap {n_i} > {n_{i + 1}}\}$ as wide shots, and the rest shots $ \{s_\mathcal{D}\}_{\mathcal{D} =[{t-{T/2}+1},{t+{T/2}}]\backslash{\mathcal{W}}}$ as detail shots.

We then use the cosine similarity between detail shots and wide shots as well as the temporal proximity between them to obtain detail-wide shots affiliation between $i \in \mathcal{D}$, $j \in \mathcal{W}$ as follow,
\begin{equation}
{j^{*}_i} = \mathop {{\mathop{\rm argmax}\nolimits} }\limits_j ({S_{ij}^{\rm{P}}} + D_{ij}),
\label{eq:affine_lg}
\end{equation}
where ${D_{ij}}= \frac{1}{{\left| {i - j} \right|}}$ measures the temporal proximity between shots $s_i$ and $s_j$.
The motivation for $D_{ij}$ is to prevent linking detail shots with wide shots that are far apart, thus ensuring that the associated wide and detail shots describe the same place.
It is worth noting that due to the inability to access the focal length of the shots, the wide shots we found by the similarity between shots are sometimes different from the wide shots in reality.

With the detail-wide shots affiliation established, we proceed to capture the short-term shot dependency based on the place features ${{\boldsymbol{X}}^{\rm{P}}}$. 
Considering the hierarchical structure of the space depicted by the wide shot and its corresponding detail shot, we devise a two-stage graph reasoning process.

On the one hand, each detail shot presents the spatial details of its corresponding wide shots. For message passing from detail shots to wide shots, we build detail-to-wide graph ${G^{{\mathcal{D}2\mathcal{W}}}} =  < {{\boldsymbol{X}}^{\rm{P}}},{{\boldsymbol{E}}^{{\rm{D2W}}}} >$. The edge weight ${\boldsymbol{E}^{\rm{D2W}}} = \{E_{ij}^{\rm{D2W}}\}$ is formulated as:
\begin{equation}
E_{ji}^{\rm{D2W}} = \left\{ {\begin{array}{*{20}{l}}
{{{({\boldsymbol{W}}_1^{\rm{D2W}}{\boldsymbol{x}}_j^{\rm{P}})}^T}{\boldsymbol{W}}_2^{\rm{D2W}}{\boldsymbol{x}}_i^{\rm{P}},}&{j = j_i^*},\\
{ - \infty ,}&{{\rm{otherwise}},}
\end{array}} \right.
\label{eq:l2g}
\end{equation}
where ${\boldsymbol{W}}_1^{{\rm{D2W}}}$ and ${\boldsymbol{W}}_2^{{\rm{D2W}}}$ are learnable matrices.

Next, we perform message passing for the shot nodes on the ${G^{{\rm{D2W}}}}$, where wide-shot nodes integrates information from its detail-shot nodes.
We apply a one-layer GCN message passing and inference to obtain short-range detail to wide shot features ${{\boldsymbol{X}}^{\rm{D2W}}}$:
\begin{equation}
{\boldsymbol{X}}^{{\rm{D2W}}} = {\rm{LN}}({{\boldsymbol{X}}^{\rm{P}}} + \sigma ({\rm{GCN}}({{\boldsymbol{X}}^{\rm{P}}},{\rm{softmax}}({{\boldsymbol{E}}^{\rm{D2W}}})))),
\label{eq:feat_lg}
\end{equation}
where we employ softmax for normalization to make the edge weights learnable~\cite{gcn_gat}.

On the other hand, we expect the updated information from wide shots can be conveyed back to the detail shots. Hence, based on short-range detail to wide shot features ${\boldsymbol{X}}^{{\rm{D2W}}}$, we build the wide-to-detail graph ${G^{{\rm{W2D}}}} =  < {{\boldsymbol{X}}^{\rm{D2W}}},{{\boldsymbol{E}}^{{\rm{W2D}}}} >$. The edge weight ${\boldsymbol{E}^{\rm{W2D}}} = \{E_{ij}^{\rm{W2D}}\}$ is formulated as:
\begin{equation}
E_{ij}^{\rm{W2D}} = \left\{ {\begin{array}{*{20}{l}}
{{{({\boldsymbol{W}}_1^{\rm{W2D}}{\boldsymbol{x}}_i^{\rm{W2D}})}^T}{\boldsymbol{W}}_2^{\rm{W2D}}{\boldsymbol{x}}_j^{\rm{D2W}},}&{j = j_i^*},\\
{ - \infty ,}&{{\rm{otherwise}}.}
\end{array}} \right.
\label{eq:g2l}
\end{equation}
By learning ${\boldsymbol{W}}_1^{\rm{W2D}}$ and ${\boldsymbol{W}}_2^{\rm{W2D}}$ on ${G^{{\rm{W2D}}}}$ with a one-layer GCN message passing and inference, we have the ultimate short-range affinity features ${{\boldsymbol{X}}^{\rm{SR}}}$:
\begin{equation}
{{\boldsymbol{X}}^{{\rm{SR}}}} = {\rm{LN}}({{\boldsymbol{X}}^{{\rm{D2W}}}} + \sigma ({\rm{GCN}}({{\boldsymbol{X}}^{{\rm{D2W}}}},{\rm{softmax}}({{\boldsymbol{E}}^{{\rm{W2D}}}})))).
\label{eq:feat_gl}
\end{equation}

\subsection{Multi-shots Comparison Detection}
While each shot enriched with diverse temporal-scale information becomes more discriminative, detecting the ending shots remains a challenge based on each shot via its own context-embedded shot feature.
This is because the ending shot implies a kind of semantic difference between its two sides.
It is essential to compare the shot semantics on both sides of the target shot to make predictions.
In pursuit of this objective, we propose Multi-shots Comparison Detection (MCD).
For shot $s_t$, we compare shot semantics on its two sides, having context similarity matrix ${{\boldsymbol{M}}}$,
\begin{equation}
{M_{ij}} = {\rm{cos}(\boldsymbol{x}}_i^{\rm{LR}},{\boldsymbol{x}}_j^{\rm{LR}}) + {\rm{cos}}({\boldsymbol{x}}_i^{\rm{SR}},{\boldsymbol{x}}_j^{\rm{SR}})
\end{equation}
where $i \in [t - T/2 + 1,t]$, $j \in [t+1, t + T/2]$.

To capture intricate patterns of semantic variation, we employ a CNN-based network on ${{\boldsymbol{M}}}$, producing the relation vector ${{\boldsymbol{r}}}$,
\begin{equation}
{{\boldsymbol{r}}} = {\rm{Flatten(CNNs}}({{\boldsymbol{M}}})),
\label{eq:cnn}
\end{equation}
where ${\rm{Flatten}(\cdot)}$ is to flatten its input matrix into a one-dimension vector, and $\rm{CNNs}(\cdot)$ denotes a 4-layer VGG~\cite{vgg} Network.

Finally, we can utilize a fullly connected MLP and a sigmoid layer on the flatten ${{\boldsymbol{R}}}$ to predict the probability ${\hat y_t}$ of the shot $s_t$ being an ending shot.

\subsection{Training and Objective Functions}
We employ two loss functions in our model training: the self-supervised loss and the supervised loss. The details of these two loss functions are as follows.

\paragraph{Self-supervised loss.} Same as previous methods~\cite{msd_bassl,msd_tran4fer}, the self-supervised loss aligns predictions with pseudo-scene boundaries using binary cross-entropy loss:
\begin{equation}
{L_f} =  - {y_t^{\rm{b}}}\log ({\hat y_t}) + (1 - y_t^{\rm{b}})\log (1 - {\hat y_t}),
\end{equation}
where ${y_t^{\rm{b}}}$ denotes the pseudo label of shot $s_t$ generated by the Modified Dynamic Warping algorithm~\cite{msd_bassl}.

\paragraph{Supervised loss.}It is an ending shot prediction loss in form of the binary cross-entropy loss:
\begin{equation}
{L_f} =  - {y_t}\log ({\hat y_t}) + (1 - {y_t})\log (1 - {\hat y_t}),
\end{equation}
where ${y_t} \in \{ 0,1\}$ denotes the ground-truth binary label of shot $t$.

Combining the aforementioned losses, our MASRC supports various learning approaches: \textbf{self-supervised learning}, \textbf{fully supervised learning}, and \textbf{self-supervised transfer learning}.
In self-supervised learning, we employ the self-supervised loss for model training. In fully supervised learning, we utilize a supervised loss. Self-supervised transfer learning involves two phases, where the self-supervised loss is applied for pre-training, followed by the use of the supervised loss for model fine-tuning.

\section{Experiments}
\subsection{Settings}
\subsubsection{Datasets:} We assess the performance of our method on three widely used video scene detection datasets,~\ie, MovieNet~\cite{msd_movienet}, BBC~\cite{msd_bbc}, and OVSD~\cite{msd_ovsd}.
\paragraph{MovieNet.} It is a vast dataset with 1,100 movies and 1.6 million shots.~318 movies are annotated with scene boundaries, forming the MovieScenes dataset~\cite{msd_lgss} for video scene detection.
MovieScenes is further divided into subsets of 190 movies for training, 64 for validation, and 64 for testing.
For different learning ways, we remain consistent with the settings of the previous methods~\cite{msd_scrl,msd_cat} and always evaluate our model on the test split of MovieScenes.
In the self-supervised scenario, we utilize the 660 unlabeled videos from MovieNet for pre-training.
In the supervised setting, we utilize 190 training videos from MovieScenes for training.
For self-supervised transfer learning, we utilize the 660 unlabeled videos from MovieNet for pre-training, and 190 training videos from MovieScenes for fine-tuning.
\paragraph{OVSD.} It consists of 21 short films, each lasting approximately 30 minutes. It contains a total of 10,000 shots and 300 scenes, extracted from movie scripts. Due to lacking predefined splits, we follow prior studies~\cite{msd_scrl,msd_bassl,msd_tran4fer}, training our model using the MovieNet dataset and then assessing its performance on OVSD without additional fine-tuning.
\paragraph{BBC.} It comprises of 11 episodes from the BBC educational TV series \textit{Planet Earth}~\cite{planetearth}. These videos have an average duration of 50 minutes and include a total of 670 scenes and 4.8K shots. As with our evaluation on the OVSD dataset, we train our model using the MovieNet dataset and assess its performance on the BBC dataset without additional fine-tuning, following established research practices~\cite{msd_scrl,msd_bassl,msd_tran4fer}.
\subsubsection{Metrics:} To measure the performances, we use the same evaluation metrics used in prior methods~\cite{msd_bassl, msd_scrl, msd_tran4fer}, which include the Average Precision (AP), the mean Intersection over Union (mIoU), and the F1-score (F1). These metrics serve to evaluate the effectiveness of video scene detection, with higher values indicating better performance.
\subsubsection{Implementation Details:}\label{subsubsec:implement}We take $T=14$ neighboring shots as input to our model.~In MASRC, we set the activation functions $\sigma(\cdot)$ in~\cref{eq:feat_lr,eq:feat_lg,eq:feat_gl} as ReLU~\cite{relu}.
In \cref{eq:vg}, we specify the the number of top similar shots as $k=4$.
For model training, we employ the Adam~\cite{adam} optimizer with a mini-batch size of 512.
For fully supervised learning and self-supervised learning, we initialize the learning rate at $10^{-4}$.~In the case of self-supervised transfer learning, we set the initial learning rate to $10^{-3}$ for pre-training and reduce it to $10^{-5}$ for fine-tuning.~Across all training stages, we apply a linear warm-up strategy during the initial epoch, followed by a learning rate decay according to a cosine schedule~\cite{cosine_learn}.
We train our MASRC on a NVIDIA RTX 3060 GPU.
In all experiments, we report the average of metrics across five different random seeds.

\subsection{Comparison with State-of-the-Art Methods}
\begin{table}[!t]
\centering
\resizebox{\columnwidth}{!}{%
\setlength{\tabcolsep}{.5mm}
\begin{tabular}{lccccc}
\hline
\multicolumn{1}{l|}{Methods}                                      & \multicolumn{1}{c|}{Modalities}              & \multicolumn{1}{c|}{Training Par.}              & AP            & mIoU          & F1            \\ \hline
\multicolumn{6}{c}{\cellcolor[HTML]{EFEFEF}Self Supervision}                                                                                                                                   \\ \hline
\multicolumn{1}{l|}{BaSSL~\cite{msd_bassl}}             & \multicolumn{1}{c|}{Entity}                  & \multicolumn{1}{c|}{43.8 M} & 31.6          & 39.4          & 32.6          \\
\multicolumn{1}{l|}{SSM~\cite{msd_ssm}}                 & \multicolumn{1}{c|}{Entity}                  & \multicolumn{1}{c|}{32.5 M} & 33.3 & 38.1          & 32.2          \\
\multicolumn{1}{l|}{TranS4mer~\cite{msd_tran4fer}}      & \multicolumn{1}{c|}{Entity}                  & \multicolumn{1}{c|}{32.0 M} & 34.5          & 39.6          & 33.4          \\
\multicolumn{1}{l|}{VSMBD~\cite{msd_vsmbd}}             & \multicolumn{1}{c|}{Entity, Place}           & \multicolumn{1}{c|}{26.5 M} & 38.3          & 42.7          & 37.9          \\
\multicolumn{1}{l|}{NeighborNet~\cite{msd_neighbor}}    & \multicolumn{1}{c|}{Entity}                  & \multicolumn{1}{c|}{35.5 M}  & 51.2          & 52.9          & 46.4          \\ \hline
\multicolumn{1}{c|}{}                                             & \multicolumn{1}{c|}{Entity}                  & \multicolumn{1}{c|}{10.7 M} & 47.3          & 48.6          & 42.0          \\
\multicolumn{1}{c|}{}                                             & \multicolumn{1}{c|}{Place}                   & \multicolumn{1}{c|}{12.9 M} & 48.9          & 47.0          & 41.9          \\
\multicolumn{1}{c|}{\multirow{-3}{*}{MASRC (Ours)}}               & \multicolumn{1}{c|}{Entity, Place}           & \multicolumn{1}{c|}{21.4 M} & \textbf{52.8} & \textbf{56.8} & \textbf{53.0} \\ \hline
\multicolumn{6}{c}{\cellcolor[HTML]{EFEFEF}Fully Supervision}                                                                                                                                  \\ \hline
\multicolumn{1}{l|}{Temporal Perceiver~\cite{msd_tp}}  & \multicolumn{1}{c|}{Entity}                  & \multicolumn{1}{c|}{52.1 M} & 53.3          & 53.2          & -             \\
\multicolumn{1}{l|}{MHRT~\cite{msd_mhrt}}               & \multicolumn{1}{c|}{Entity, Place, Audio}    & \multicolumn{1}{c|}{47.1 M} & 54.8          & 51.2          & 46.3          \\
\multicolumn{1}{l|}{CANet~\cite{msd_canet}}             & \multicolumn{1}{c|}{Face, Body}              & \multicolumn{1}{c|}{15.3 M} & 56.8          & 55.7          & -             \\
\multicolumn{1}{l|}{NeighborNet~\cite{msd_neighbor}}    & \multicolumn{1}{c|}{Entity}                  & \multicolumn{1}{c|}{35.5 M}  & 64.0          & 61.2          & 57.8          \\ \hline
\multicolumn{1}{c|}{}                                             & \multicolumn{1}{c|}{Entity}                  & \multicolumn{1}{c|}{10.7 M} & 59.2          & 58.5          & 55.0          \\
\multicolumn{1}{c|}{}                                             & \multicolumn{1}{c|}{Place}                   & \multicolumn{1}{c|}{12.9 M} & 59.6          & 57.9          & 52.9          \\
\multicolumn{1}{c|}{\multirow{-3}{*}{MASRC (Ours)}}               & \multicolumn{1}{c|}{Entity, Place}           & \multicolumn{1}{c|}{21.4 M} & \textbf{67.4} & \textbf{65.8} & \textbf{63.8} \\ \hline
\multicolumn{6}{c}{\cellcolor[HTML]{EFEFEF}Self-Supervised Transfer}                                                                                                                           \\ \hline
\multicolumn{1}{l|}{ShotCoL~\cite{msd_shotcol}}         & \multicolumn{1}{c|}{Entity}                  & \multicolumn{1}{c|}{26.3 M} & 53.4          & -             & 51.4          \\
\multicolumn{1}{l|}{SCRL~\cite{msd_scrl}}               & \multicolumn{1}{c|}{Entity}                  & \multicolumn{1}{c|}{26.3 M} & 54.6          & -             & 51.4          \\
\multicolumn{1}{l|}{BaSSL~\cite{msd_bassl}}             & \multicolumn{1}{c|}{Entity}                  & \multicolumn{1}{c|}{43.8 M} & 57.4          & 50.7          & 47.0          \\
\multicolumn{1}{l|}{SSM~\cite{msd_ssm}}                 & \multicolumn{1}{c|}{Entity}                  & \multicolumn{1}{c|}{32.5 M} & 59.7          & 51.3          & 48.4          \\
\multicolumn{1}{l|}{CAT~\cite{msd_cat}}                 & \multicolumn{1}{c|}{Entity}                  & \multicolumn{1}{c|}{43.8 M} & 59.6          & 53.7          & 51.9          \\
\multicolumn{1}{l|}{TranS4mer~\cite{msd_tran4fer}}      & \multicolumn{1}{c|}{Entity}                  & \multicolumn{1}{c|}{32.0 M} & 60.8          & 51.9          & 48.4          \\
\multicolumn{1}{l|}{VSMBD~\cite{msd_vsmbd}}             & \multicolumn{1}{c|}{Entity, Place}           & \multicolumn{1}{c|}{26.5 M} & 63.7          & 56.4          & 55.3          \\
\multicolumn{1}{l|}{NeighborNet~\cite{msd_neighbor}}    & \multicolumn{1}{c|}{Entity}                  & \multicolumn{1}{c|}{35.5 M} & 71.9 & 64.5 & 62.7 \\ \hline
\multicolumn{1}{c|}{}                                             & \multicolumn{1}{c|}{Entity}                  & \multicolumn{1}{c|}{10.7 M} & 65.5          & 62.6          & 64.0          \\
\multicolumn{1}{c|}{}                                             & \multicolumn{1}{c|}{Place}                   & \multicolumn{1}{c|}{12.9 M} & 65.9          & 61.4          & 61.3          \\
\multicolumn{1}{c|}{\multirow{-3}{*}{MASRC (Ours)}}               & \multicolumn{1}{c|}{Entity, Place}           & \multicolumn{1}{c|}{21.4 M} & \textbf{73.2} & \textbf{70.7}  & \textbf{67.3}    \\ \hline
\end{tabular}%
}
\caption{Performance comparison between our proposed method and recent baselines on the MovieNet dataset~\cite{msd_movienet}. The best results for each category of methods are indicated in \textbf{bold}.}
\label{tab:sota_movinet}
\end{table}

\begin{table}[!t]
\centering
\resizebox{.8\columnwidth}{!}{%
  \begin{tabular}{l|cc}
    \hline
    Methods & OVSD & BBC \\ \hline
    {ShotCoL~\cite{msd_shotcol}} & 25.5 & 28.0 \\
    {SCRL~\cite{msd_scrl}} & 38.8 & 30.2 \\
    {BaSSL~\cite{msd_bassl}} & {28.7} & {40.0} \\
    {TranS4mer~\cite{msd_tran4fer}} & {36.0} & {43.6} \\
    {NeighborNet~\cite{msd_neighbor}} & 47.3 & {50.6}\\
    {MASRC} & \textbf{48.3} & \textbf{53.2} \\
    \hline
  \end{tabular}}
\captionof{table}{Cross dataset transfer result (AP) on OVSD~\cite{msd_ovsd} and BBC~\cite{msd_bbc} without fine-tuning.} 
\label{tab:sota_transfer}
\end{table}
\subsubsection{Results on MovieNet.}\Cref{tab:sota_movinet} displays the quantitative results on MovieNet~\cite{msd_movienet}.
Given that numerous methods rely solely on single-modal features, we showcase performance of our method in the corresponding modality for a fair comparison.
Results show that, in contrast to counterparts, our method consistently excels with fewer training parameters across various experimental scenarios.
Compared with multi-modal methods, our method employs less modality knowledge and outperforms the state-of-the-art MHRT~\cite{msd_mhrt}.
The superior performance of our approach arises from diversely modeling inter-shot relations and detecting scenes through comparisons between multi-shots relations.
\subsubsection{Transfer Evaluation.}We demonstrate the generalization capability of our MASRC in comparison with recent methods~\cite{msd_shotcol,msd_scrl,msd_bassl,msd_tran4fer} in~\Cref{tab:sota_transfer}.~All models used have undergone self-supervised pre-training and fine-tuning on MovieNet~\cite{msd_movienet}.~We test these MovieNet-trained model without any additional fine-tuning on BBC~\cite{msd_bbc} and OVSD~\cite{msd_ovsd}.~The results demonstrate that our proposed method achieves the best performance among all the comparisons on the OVSD and BBC datasets, which verifies the strong generalization capability of MASRC.
\subsection{Ablation Studies}
The following results are obtained in a fully supervised manner. We have also included more ablation experiments and visualizations in the supplementary.
\begin{figure}[!t]
\centering
\begin{minipage}[t]{0.22\textwidth}
\centering
\includegraphics[width=\textwidth]{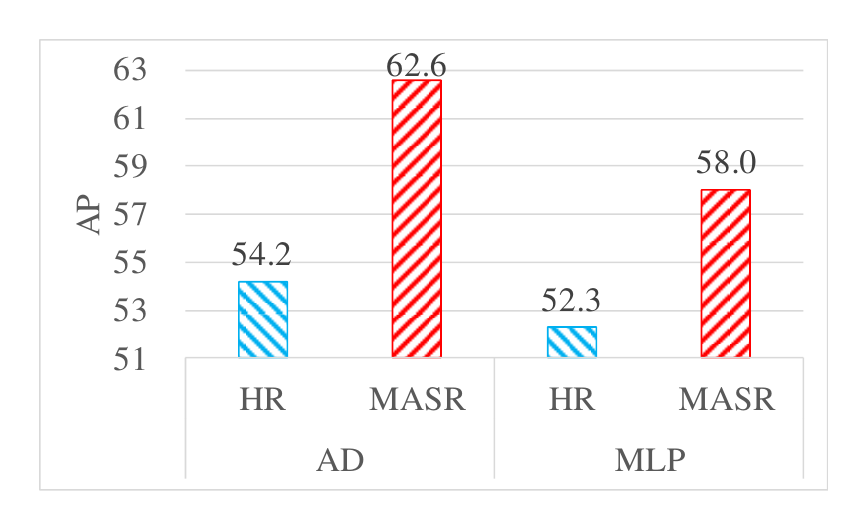}
\caption{\small{Comparison between our MASR and HR~\cite{msd_mhrt} under the same detector, AD~\cite{msd_mhrt} or MLP~\cite{msd_tran4fer}.}}
\label{fig:abl_masr_general}
\end{minipage}
{ }
\begin{minipage}[t]{0.22\textwidth}
\centering
\includegraphics[width=\textwidth]{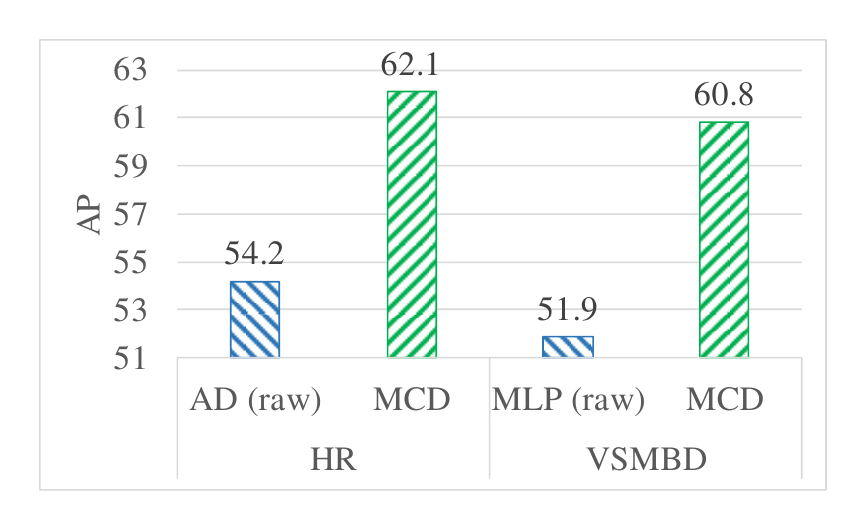}
\caption{\small{Comparison of our MCD with their raw detectors on temporal modeling methods, HR~\cite{msd_mhrt} and VSMBD~\cite{msd_vsmbd}.}}
\label{fig:abl_mcd_general}
\end{minipage}

\begin{minipage}[t]{0.22\textwidth}
\centering
\includegraphics[width=\textwidth]{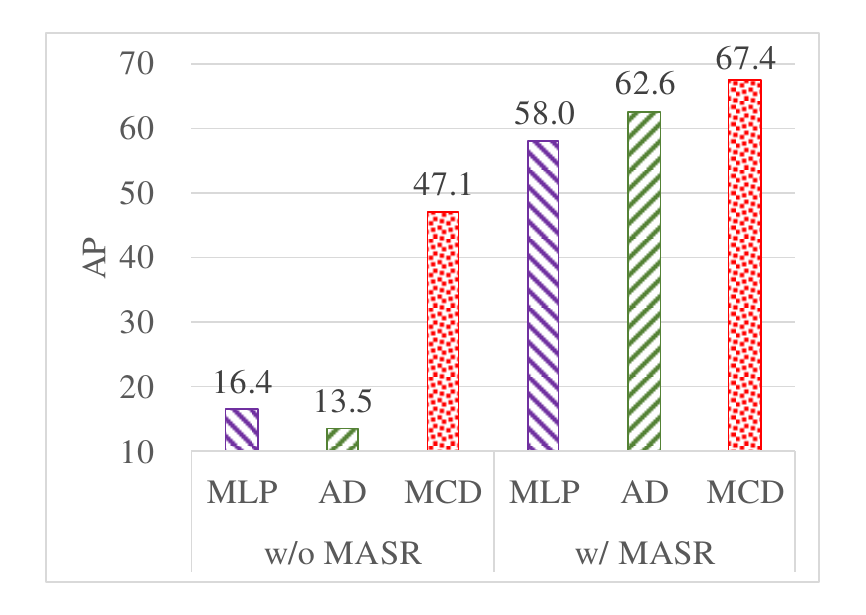}
\caption{\small{Comparison of detectors, including MLP~\cite{msd_tran4fer}, AD~\cite{msd_mhrt}, and our MCD.}}
\label{fig:abl_detector}
\end{minipage}
{ }
\begin{minipage}[t]{0.22\textwidth}
\centering
\includegraphics[width=\textwidth]{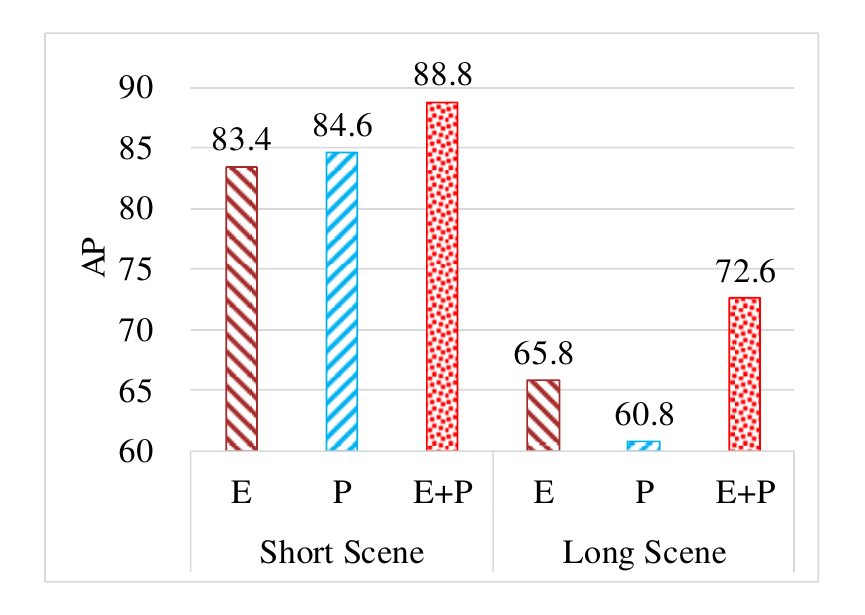}
\caption{\small{Influence of Entity (E) and Place (P) modalities on detection in video scenes with varied lengths.}}
\label{fig:abl_modal_scene}
\end{minipage}
\end{figure}
\subsubsection{Pluggability of MASR.}\cref{fig:abl_masr_general} provides the performance of HR \cite{msd_mhrt} and our MASR combined with different ending shot detectors, including MLP~\cite{msd_bassl,msd_cat,msd_tran4fer} and AD~\cite{msd_mhrt}.
The results show that combined with the same detector, the performance of our proposed MASR is better than HR, highlighting the contribution and plug-and-play ability of our proposed MASR.
\subsubsection{Pluggability of MCD.}\cref{fig:abl_mcd_general} shows the performance of our proposed MCD combined with different temporal relation modeling methods, including HR~\cite{msd_mhrt} and VSMBD~\cite{msd_vsmbd}.
The results highlight that the combination of our proposed MCD with different methods is better than their raw alternatives, highlighting the advantage and versatility of MCD.

\subsubsection{Different Ending Shot Detectors.}\cref{fig:abl_detector} compares results about different detectors, where MLP~\cite{msd_bassl,msd_cat,msd_tran4fer}and AD~\cite{msd_mhrt} are two competitors of our MCD detector.
The former makes predictions through single-shot representation, and the latter uses learnable class vectors to query each shot feature to decide whether each shot is a boundary.
Our proposed MCD outperforms the competitors under evaluation metrics, mainly because MCD explicitly uses shot contexts to predict boundaries.

\subsubsection{Impact of Each Modality on Detecting Video Scenes of Different Lengths.}\cref{fig:abl_modal_scene} provides an in-depth analysis of the impact of different modal inputs on video scene detection. Short scenes comprise fewer than 12 shots, while long scenes consist of a higher number of shots, based on the fact that the average number of scenes in MovieNet~\cite{msd_lgss} is 12.6. As can be seen from \cref{fig:abl_modal_scene}, compared to the entity modality, the place modality shows a greater advantage in short scene detection. Conversely, the entity modality outperforms the place modality in detecting long scenes. This observation validates that the place modality is well-suited for capturing short-term shot relations, whereas the entity modality excels in capturing long-range shot relations. Notably, combined modalities consistently perform optimally under either scene length, confirming the complementarity of the two modalities.
\begin{table}[!t]	
\centering
\resizebox{0.97\columnwidth}{!}{%
\begin{tabular}{cccc|l}
\hline
\multicolumn{2}{c|}{Entity}      & \multicolumn{2}{c|}{Place} & \multicolumn{1}{c}{\multirow{2}{*}{AP}} \\ \cline{1-4}
Short Term         & \multicolumn{1}{c|}{Long Term} & Short Term           & Long Term          & \multicolumn{1}{c}{}  \\ \hline
\checkmark &       -                 &   -          &  -          & 56.0 \\
   -       & \checkmark              &   -          &  -          & 59.2 \\
   -       &     -                   &   -          & \checkmark  & 57.3 \\
   -       &     -                   & \checkmark   &     -       & 59.6 \\
\checkmark &      -                  &      -       & \checkmark  & 63.2 \\
     -     & \checkmark              & \checkmark   &     -       & 67.4 \\ \hline
\end{tabular}%
}
\caption{Comparison of different combinations of modalities and shot relations modeling.}
\label{tab:val_lsap}
\end{table}	
\subsubsection{Each Modality and Its Optimal Modeling Temporal Relations.}\Cref{tab:val_lsap} displays the performance of temporal relation modules with inputs from various modalities.
The long-term relations module is modeled by \cref{eq:vg,eq:feat_lr}, and the short-term relations module is built upon Eqs.~\eqref{eq:threhold}-\eqref{eq:feat_gl}.
For the entity modality, switching from short-term to long-term relations modeling improves AP by 3.2\%. Conversely, for the place modality, switching to short-term relations modeling improves AP by 3.6\%.
The best performance is achieved by combining long-term entity modeling with short-term place modeling.
In summary, the entity modality is suited for modeling long-term shot relations, whereas the place modality is adept at capturing short-term shot relations.

\begin{table}[!t]
\centering
\resizebox{.5\linewidth}{!}{%
\begin{tabular}{ccc|c}
\hline
ELD        & PSD        & MCD        & AP \\ \hline
\textbf{-} & \textbf{-} & \textbf{-} & 16.4 \\
-          & -          & \checkmark & 47.1\\
\checkmark & -          & \checkmark & 59.2\\
-          & \checkmark & \checkmark & 59.6 \\
\checkmark & \checkmark & \checkmark & 67.4 \\ \hline
\end{tabular}%
}
\caption{Ablation study on MASRC in different inclusions of entity-based long-term dependency (ELD), place-based short-term dependency (PSD), and multi-shots comparison detection (MCD). \checkmark signifies ``included'', while - ``excluded''.}
\label{tab:abl_component}
\end{table}

\subsubsection{Network Components.}\Cref{tab:abl_component} presents the results of assessing the contributions of different components of our proposed method to overall performance.
The first row reports baseline results of feeding the concatenation of entity and place features into the MLP to predict shot classes.
Replacing the MLP detector with the proposed MCD results in a significant improvement across all metrics, emphasizing the effects of contrasting shot contexts to detect ending shots.
The addition of the ELD or PSD module alone leads to increment in AP, which arises from our consideration of the distinct roles of various visual semantics in video scene detection.
The combination of ELD and PSD achieves peak performance, demonstrating their complementary effects.

\subsubsection{Alternatives to CNNs in MCD.} \Cref{tab:abl_cnns} presents performance of alternatives to CNNs in MCD. As shown, CNN-based models, such as UnirepkNet~\cite{unireplknet} and our used CNNs in \cref{eq:cnn}, achieve better results compared to other alternatives, likely because CNN models are better suited to capturing the variation patterns of gridded data.
\begin{table}[t]
\centering
\resizebox{\linewidth}{!}{%
\begin{tabular}{l|cccc}
\hline
Alternatives & Max Pooling & Self Attention & CNNs (used) & Unireplknet \\ \hline
AP      & 56.7        & 59.9           & 67.4            & 68.9        \\ \hline
\end{tabular}%
}
\caption{Comparison of alternatives to our used CNNs.}
\label{tab:abl_cnns}
\end{table}

\begin{figure}[!t]
	\centering
	\subfloat[\small Temporal window scale]{
	  \includegraphics[width=0.47\linewidth]{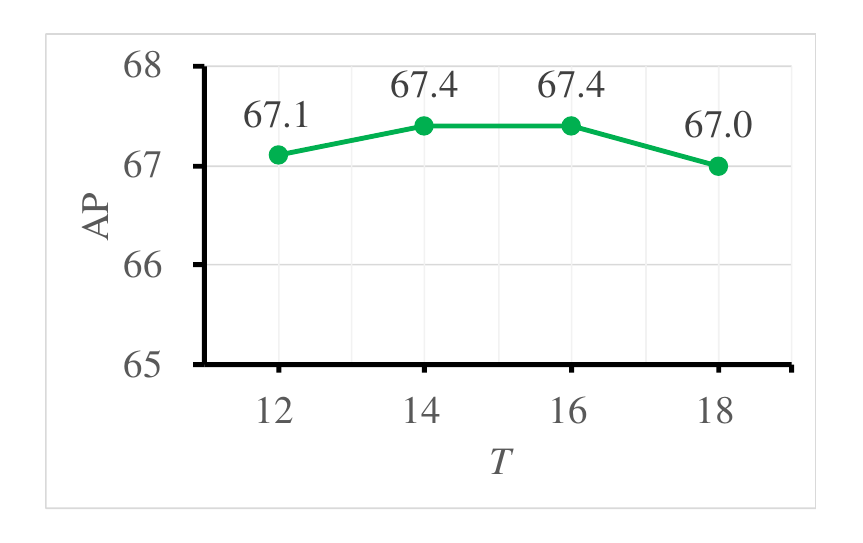}
	  \label{fig:abl_time_slide}
	}
	\subfloat[\small Entity feature neighbor scale]{
		\includegraphics[width=0.47\linewidth]{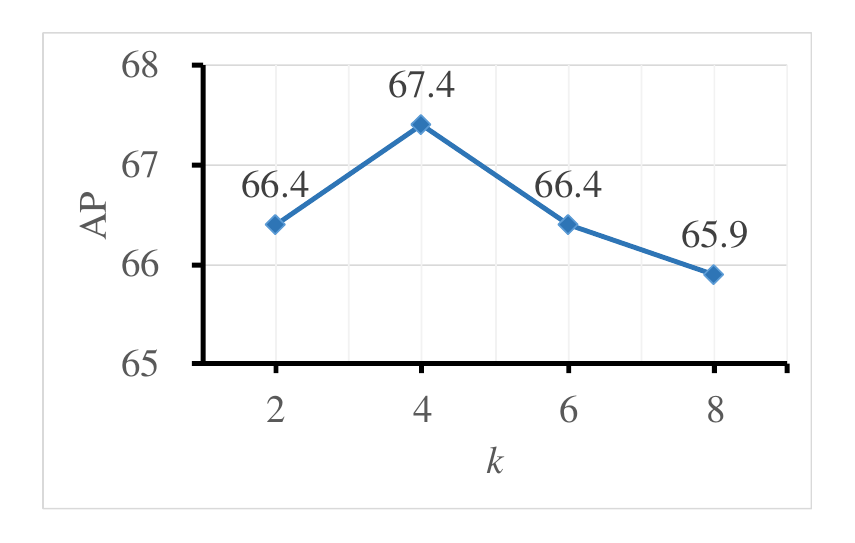}
		\label{fig:abl_topk}
		}
	\caption{Impact of hyperparameters. (a) Temporal window scale $T$ as defined in \cref{subsec:mldm}. (b) Entity feature neighbor scale $k$ as defined by top-$k$ in \cref{eq:vg}.}
\end{figure}

\subsubsection{Temporal Window Scale $T$.}\cref{fig:abl_time_slide} illustrates the impact of the time sliding window scale $T$, as defined in \cref{subsec:mldm}. It is evident that all metrics reach their peaks when $T=14$. This observation aligns with the average number of shots per scene in the MovieNet dataset which is 12.6.

\subsubsection{Entity Feature Neighbor Scale $k$.}\cref{fig:abl_topk} presents the effects of the amount of most similar shots $k$ defined in \cref{eq:vg}. The proposed method achieves optimal performance when $k=4$. This selection strikes the best trade-off between capturing sufficient similar entity variations and avoiding the introduction of extraneous noisy information.

\section{Conclusion}
We propose an novel multi-modal shot relationship modeling and comparison framework for video scene detection.
It reasons on the established long-range entity dependency graph and short-range place dependency graph to capture long-short dependencies between shots.
For detecting ending shots, predictions are made by comparing the semantic relationships of surrounding shots to the target shot.
Experimental results in public datasets show the effectiveness and superiority of our MASRC over previous SOTAs.

\section*{Acknowledgements}
This work is supported in part by the National Natural Science Foundation of China under Grant
61976029 and the Key Project of Chongqing Technology Innovation and Application Development under Grant cstc2021jscx-gksbX0033.

\bibliography{refs}

\newpage

\clearpage
\newcounter{counter}[section]
\twocolumn[{%
\renewcommand\twocolumn[1][]{#1}%
\begin{center}
    \Large
    \textbf{Supplementary Material for\\ Modality-Aware Shot Relating and Comparing for Video Scene Detection}
    \\[20pt]

\end{center}
}]

The following results are obtained in a fully supervised manner on the MovieNet~\cite{msd_movienet} dataset, unless otherwise specified.
\section{Additional Experimental Results}
\label{sec:other_abl}
\subsection{Ablation Studies on Place Continuity Graph}
\begin{figure}[!h]
\centering
\includegraphics[width=\linewidth]{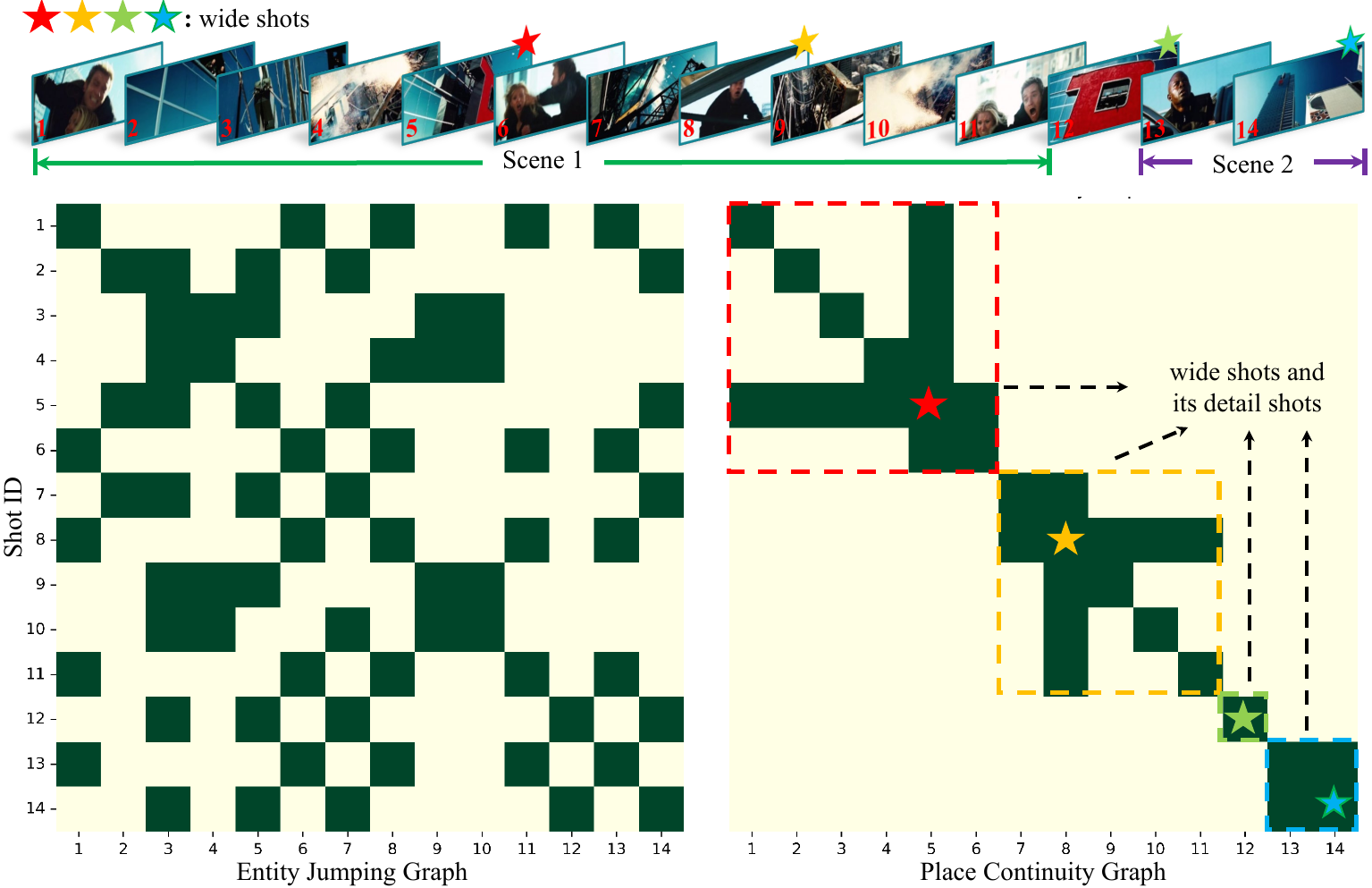}
\caption{Visualization of edges connections (in dark color) of EJG and PCG. Stars indicate different wide shots discovered by the place modality, and the others are detail shots.}
\label{fig:vis_ejg_cpg}
\end{figure}

\paragraph{Detail-Wide Shot Affiliation.}The composition of detail-wide shot affiliation, as defined in Eq. (4), is ablated in \Cref{tab:affine_lg}. The results indicate that combining shot similarity and temporal proximity yields optimal performance. This underscores the importance of considering both shot similarity and temporal proximity when determining the affiliation between wide shots and detail shots.
\begin{table}[!h]
\centering
\begin{tabular}{l|ccc}
\hline
\multicolumn{1}{c|}{Local-Global Shot Affiliation} & AP   & mIoU & F1   \\ \hline
Shot Similarity  $S_{ij}^{\rm{P}}$                          & 65.2 & 64.7 & 63.3 \\
Temporal Proximity $D_{ij}$                           & 63.9 & 60.1 & 59.9 \\
Both                                                & 67.4 & 65.8 & 63.8 \\ \hline
\end{tabular}%
\caption{Comparison of variants of local-global shots affiliation as defined in Eq. (4).}
\label{tab:affine_lg}
\end{table}
\paragraph{Necessary of Using Both Detail-to-Wide (D2W) and Wide-to-Detail (W2D) Graphs.}\Cref{tab:abl_d2w_w2d} presents the results of assessing the contributions of different components of our proposed PCG to overall performance.  Combining D2W and W2D achieves significantly better performance compared to using only D2W or only W2D.
\begin{table}[!h]
\centering
\begin{tabular}{cc|c}
\hline
D2W        & W2D        & AP   \\ \hline
\checkmark & -          & 63.6 \\
-          & \checkmark & 65.3 \\
\checkmark & \checkmark & 67.4 \\ \hline
\end{tabular}
\caption{Ablation study on PCG graph in different inclusions of D2W and W2D graphs. \checkmark signifies ``included", while - ``excluded''.}
\label{tab:abl_d2w_w2d}
\end{table}
\paragraph{Visualization of Edge Connections in EJG and PCG.}In \cref{fig:vis_ejg_cpg}, we present a sample result of edge connections of EJG and PCG, built upon Sec. 3.2. 
As expected, EJG connects similar but distant shots, while PCG links consecutive shots depicting the same location.
Consequently, EJG is advantageous for modeling long-range shot relations, whereas PCG is effective for modeling short-range shot relations.
\subsection{Comparison with MHRT under Various Modality Combinations}
\begin{figure}[!t]
	\centering
	  \includegraphics[width=0.8\linewidth]{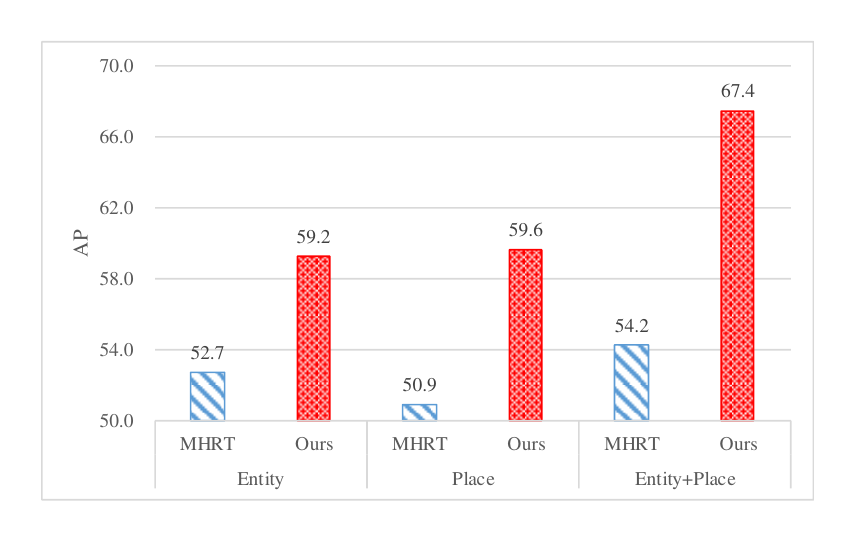}
	  \caption{Comparison between proposed MASRC and MHRT~\cite{msd_mhrt} under various combinations of multi-modality semantic inputs.}
	  \label{fig:abl_modal}
\end{figure}

\cref{fig:abl_modal} compares the performance of our MASRC with MHRT\footnote{The code for MHRT is not available; hence, we reproduced it for comparisons.}~\cite{msd_mhrt}, a state-of-the-art multi-modal video scene detection method, under the same entity and place feature inputs. Our model beats MHRT regardless of the combination of different visual semantic modalities. This superiority of ours is mainly thanks to diverse shot relationship modeling and context comparison convolution for ending shot detection.
\subsection{Computation Cost Analysis} Under identical hardware conditions, \Cref{tab:speed} presents the number of training parameters, the training/inference throughput measured in samples per second (Sam./s), and GPU memory costs per sample (GB/Sam.) for state-of-the-art methods. A sample corresponds to a time window with a length of 14 shots, equating to 14 nodes in our graph.~As can be seen, our model requires fewer resources in terms of training params, training/inference throughput, and GPU memory costs compared to others utilizing the same backbone.
\begin{figure*}[!t]
\centering
\includegraphics[width=.9\linewidth]{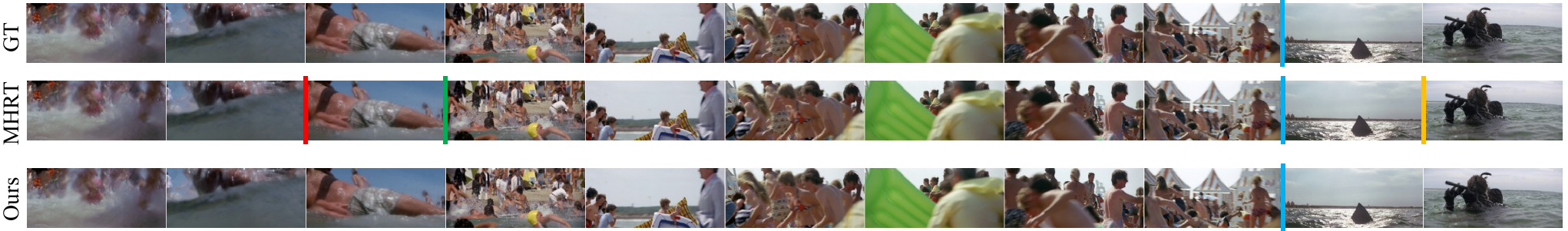}
\caption{Qualitative comparison of the proposed method with the previous method MHRT~\cite{msd_mhrt}. GT denotes ground-truth scene boundaries for reference. The colored vertical lines represent the video scene boundaries.}
\label{fig:vis_short_detect}
\end{figure*}
\begin{figure*}[!t]
\centering
\includegraphics[width=\linewidth]{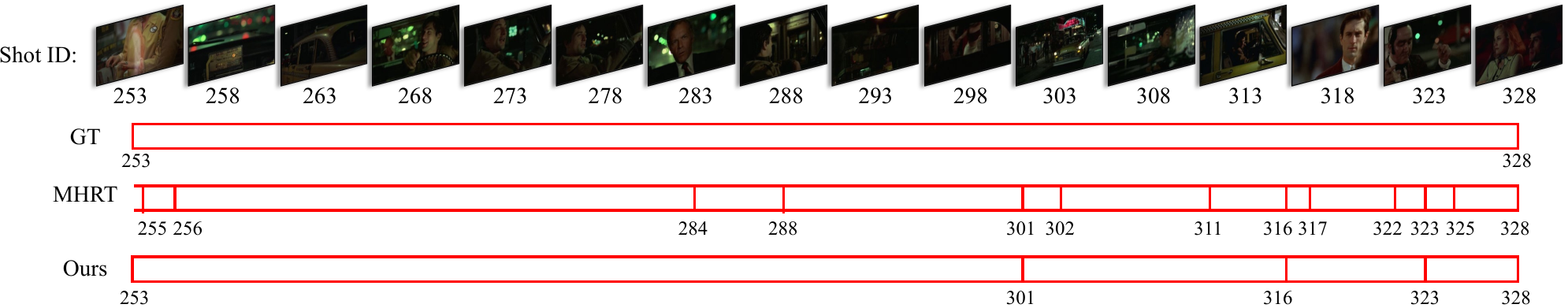}
\caption{Qualitative comparison of the proposed method with the previous method MHRT~\cite{msd_mhrt}. GT denotes ground-truth scene boundaries for reference. The red vertical lines represent the video scene boundaries.}
\label{fig:vis_long_detect}
\end{figure*}
\subsection{Visualization of Short Scene Detection.} In \cref{fig:vis_short_detect}, we present a sample results of video scene detection on the MovieNet dataset. Ground-truth annotations are provided for reference, and we include the previous state-of-the-art method, MHRT~\cite{msd_mhrt}, for comparison. The results demonstrate that MHRT is susceptible to camera shot changes, including switching, zooming, and movement, leading to over-segmented video scenes. In contrast, our proposed method effectively handles these challenging cases without experiencing over-segmentation. This is attributed to our approach on establishing the diverse relations between shots and carefully comparing the relations between shots.
\begin{table}[!t]
\centering
\resizebox{\columnwidth}{!}{%
\begin{tabular}{l|c|c|cccc}
\hline
\multicolumn{1}{c|}{\multirow{2}{*}{Methods}} & \multirow{2}{*}{Modalities} &\multirow{2}{*}{Train Par.} & \multicolumn{2}{c|}{Train}   & \multicolumn{2}{c}{Inference} \\ \cline{4-7} 
\multicolumn{1}{c|}{}                         &                &             & Sam./s    & \multicolumn{1}{c|}{GB/sam.} & Sam./s        & GB/sam.       \\ \hline
LGSS~\cite{msd_lgss}                                          & P              & 13.7 M             & 32     & 0.37                         & 68            & 0.17          \\
Ours                                          & P              & 12.9 M             & 3788    & 0.0013                       & 5632          & 0.0005        \\
MHRT~\cite{msd_mhrt}                                         & E+P             & 27.8 M             & 2509    & 0.0020                       & 3584          & 0.0014        \\
Ours                                          & E+P            & 21.4 M             & 3093    & 0.0019                       & 5171          & 0.0009        \\ \hline
\end{tabular}
}
\caption{Number of training parameters (Train Par.), training/inference throughput measured in samples per second (Sam./s), and GPU memory costs per sample (GB/Sam.) for state-of-the-art methods. E is short for entity modality and P is for place modality.}
\label{tab:speed}
\end{table}

\subsection{Visualization of Long Scene Detection.} \cref{fig:vis_long_detect} depicts a result of long scene consisting of 76 shots. It portrays a taxi driver picking up two separate groups of passengers during his night shift before finishing his work. The sequence involves numerous rapid shot transitions. Previous SOTA~\cite{msd_mhrt} incorrectly segments the scene into 12 scenes, while our method reduces false positive detection and segments the scene into 5 scenes. This is attributed to our approach on establishing modality-aware shot relations and comparing the relations between shots.
\subsection{Limitation.} 
\begin{figure}[!t]
\centering
\includegraphics[width=\linewidth]{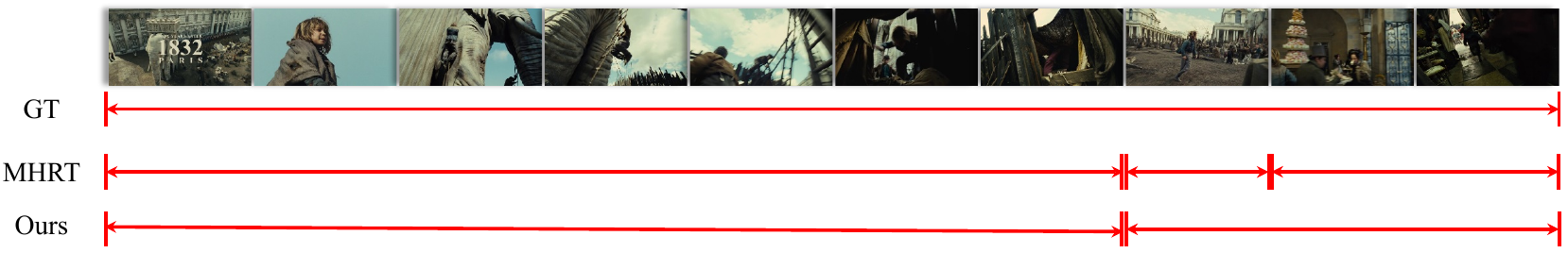}
\caption{Qualitative comparison of the proposed method with the previous method MHRT~\cite{msd_mhrt}. GT denotes ground-truth scene boundaries for reference. The red vertical lines represent the video scene boundaries.}
\label{fig:failure}
\end{figure}

We find some commonalities in cases where our method and previous SOTA method~\cite{msd_mhrt} may fail. \cref{fig:failure} presents a short scene from "Les Misérables". Both our method and previous SOTA~\cite{msd_mhrt} segment this scene into two parts. The inconsistency between shots arises from frequent motion transitions, which weaken the coherence of entities and places, leading to incorrect segmentation. Addressing this challenge may involve incorporating background audio, such as music, which can provide another type of contextual consistency. Exploring the integration of audio into our model is a future research direction.
\begin{figure}
\centering
\includegraphics[width=.75\linewidth]{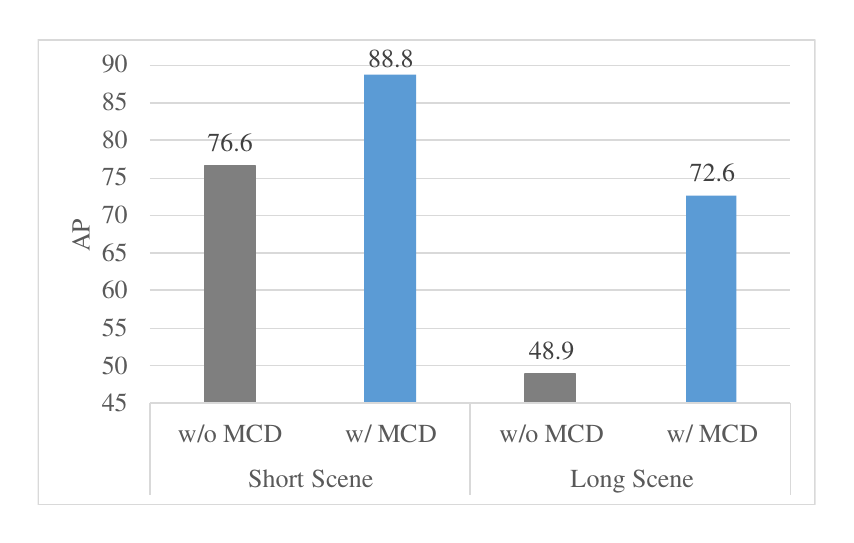}
\caption{Influence of w/o and w/ MCD on detecting video scenes with varied lengths.}
\label{fig:abl_mcd_scene}
\end{figure}
\subsection{Impact of MCD on Detecting Video Scenes of Different Lengths}\cref{fig:abl_mcd_scene} provides a deeper analysis of the impact of the proposed MCD on video scene detection. We let length 12 to distinguish between short scenes (each of which has $\le$ 12 shots) and long scenes (each of which has $>$ 12 shots) based on the fact that the average shot number in MovieNet scenes is 12.6. As shown in \cref{fig:abl_mcd_scene}, compared with the model without MCD, the model with MCD improves both short and long scene detection. Notably, in long scene detection, the inclusion of MCD (72.6\% AP) significantly enhances the performance compared to the exclusion of MCD.


\end{document}